\documentclass[11pt,a4paper]{article}
\usepackage[hyperref]{acl2020}
\usepackage{times}
\usepackage{latexsym}

\usepackage{graphicx}
\usepackage{url}

\usepackage{amsmath}
\usepackage{amsfonts,bm}

\def\eqref#1{equation~\ref{#1}}

\def\1{\bm{1}}

\def\va{{\bm{a}}}

\def\vf{{\bm{f}}}
\def\vg{{\bm{g}}}

\def\vk{{\bm{k}}}

\def\vm{{\bm{m}}}

\def\vq{{\bm{q}}}

\def\vw{{\bm{w}}}

\def\m1{{\bm{1}}}
\def\mA{{\bm{A}}}

\def\mJ{{\bm{J}}}
\def\mK{{\bm{K}}}
\def\mL{{\bm{L}}}
\def\mM{{\bm{M}}}

\def\mQ{{\bm{Q}}}

\def\mV{{\bm{V}}}
\def\mW{{\bm{W}}}

\DeclareMathAlphabet{\mathsfit}{\encodingdefault}{\sfdefault}{m}{sl}
\SetMathAlphabet{\mathsfit}{bold}{\encodingdefault}{\sfdefault}{bx}{n}

\newcommand{\softmax}{\mathcal{S}}

\DeclareMathOperator*{\argmax}{arg\,max}

\DeclareMathOperator{\real}{\rm I\!R}

\usepackage{xspace}

\newcommand{\Ni}{({\em i})~}
\newcommand{\Nii}{({\em ii})~}
\newcommand{\Niii}{({\em iii})~}
\newcommand{\Niv}{({\em iv})~}

\usepackage{tikz}
\usepackage{caption}
\usepackage{subcaption}
\usepackage{booktabs}
\usepackage{comment}
\usepackage{microtype}

\aclfinalcopy 
\def\aclpaperid{655} 

\title{Differentiable Window for Dynamic Local Attention}

\author{Thanh-Tung Nguyen$^*{\dagger}$$^\P$, Xuan-Phi Nguyen\thanks{*Equal contributions} ${\dagger}$$^\P$, Shafiq Joty$^\P$$^\S$, Xiaoli Li$^\dagger$\\
  $^\P$Nanyang Technological University \\
  $^\S$Salesforce Research Asia \\
  $^\dagger$Institute for Infocomm Research, A-STAR \\
  Singapore \\
  \texttt{\{ng0155ng@e.;nguyenxu002@e.;srjoty@\}ntu.edu.sg} 
  \\
  \texttt{xlli@i2r.a-star.edu.sg} 
}

\date{}

\begin{document}
\maketitle
\begin{abstract}
We propose Differentiable Window, a new neural module and general purpose component for dynamic window selection. While universally applicable, we demonstrate a compelling use case of utilizing Differentiable Window to improve standard attention modules by enabling  more focused attentions over the input regions. We propose two variants of Differentiable Window, and integrate them within the Transformer architecture in two novel ways. We evaluate our proposed approach on a myriad of NLP tasks, including machine translation, sentiment analysis, subject-verb agreement and language modeling. Our experimental results demonstrate consistent and sizable improvements across all tasks.
\end{abstract}

\section{Introduction} \label{sec:intro}

Computing relative importance across a series of inputs can be regarded as one of the important advances in modern deep learning research. This paradigm, commonly known as \emph{attention} \cite{bahdanau2014neural}, has demonstrated immense success across a wide spectrum of applications. To this end, learning to compute contextual representations \cite{vaswani2017attention}, to point to the relevant part in the input \cite{pointer-net}, or to select windows or spans \cite{wang2016machine} from sequences forms the crux of many modern deep neural architectures. 

Despite aggressive advances in developing neural modules for computing relative relevance \cite{luong2015effective,Chiu-18}, there has been no general purpose {solution} for learning differentiable attention windows. While span selection-based pointer network models typically predict a start boundary and an end boundary \cite{wang2016machine,seo2016bidirectional}, these soft predictions generally reside at the last layer of the network and are softly optimized. To the best of our knowledge, there exists no general purpose component for learning differentiable windows within networks.

Although the practical {advantages of learning differentiable windows are plenty, this paper focuses on improving attentions with differentiable windows.} The key idea is to enable more \textit{focused} attention, leveraging dynamic window selection for limiting (and guiding) the search space for the standard attention modules to work within. This can also be interpreted as performing a form of dynamic local attention.


We make several key technical contributions. First, we formulate the dynamic window selection problem as a problem of learning a \emph{discrete} mask (i.e., binary values representing the window). By learning and composing left and right boundaries, we show that we are able to parameterize the (discrete) masking method. We then propose \emph{soft} adaptations of the above mentioned, namely \textbf{trainable soft masking} and \textbf{segment-based soft masking}, which are differentiable approximations  {that} can not only be easily optimized in an end-to-end fashion, but also inherit the desirable properties of discrete masking.

While these modules are task and model agnostic, we imbue the state-of-the-art Transformer \cite{vaswani2017attention} model with our differentiable window-based attention. To this end, we propose two further variants, i.e., \textbf{multiplicative window attention} and \textbf{additive window attention} for improving the Transformer model. Within the context of sequence transduction and self-attention based encoding, learning dynamic attention windows are beneficial because they can potentially eliminate noisy aggregation and alignment from large input sequences. 
 
On the other hand, it is good to note that {hard attention \cite{Xu2015attention}, which replaces the weight average of soft attention with a stochastic sampling model}, tries to achieve similar ends, albeit restricted to token-level selection. Hence, our proposed differentiable windows are more flexible and expressive compared to hard attentions.

We evaluate our Transformer model with differentiable window-based attention on a potpourri of NLP tasks, namely \emph{machine translation, sentiment analysis, language modeling,} and \emph{subject-verb agreement}. Extensive experimental results on these tasks demonstrate the effectiveness of our proposed  method. Notably, on the English-German and English-French WMT'14 translation tasks, our method accomplishes improvements of 0.63 and 0.85 BLEU, respectively. On the Stanford Sentiment Treebank and IMDB sentiment analysis tasks, our approach achieves 2.4\% and 3.37\% improvements in accuracy, respectively. We further report improvements of 0.92\% in accuracy and 2.13 points in perplexity on the subject-verb agreement and language modeling tasks, respectively. We make our code publicly available at \url{https://ntunlpsg.github.io/project/dynamic-attention/}.

\section{Background} \label{sec:background}

The \emph{attention mechanism} enables dynamic selection of relevant contextual representations with respect to a query representation. It has become a key module in most deep learning models for  language and image processing tasks, especially in encoder-decoder models \cite{bahdanau2014neural,luong2015effective,pmlr-v37-xuc15}.


\subsection{Transformer and Global Attention} 

The Transformer network \cite{vaswani2017attention} models the encoding and decoding processes using stacked self-attentions and cross-attention (encoder-decoder attentions). Each attention layer uses a scaled multiplicative formulation defined as:
\begin{align}
\hspace{-1em}\text{score}(\mQ,\mK) = \frac{(\mQ\mW^Q) (\mK\mW^K)^T}{\sqrt{d}} \label{eq:score}
\end{align}

\begin{align}
\hspace{-1em}\text{att}(\mQ,\mK, \mV) = \softmax (\text{score}(\mQ,\mK)) (\mV\mW^V) 
\label{eq:att}
\end{align}

\noindent where $\softmax(\mA)$ denotes the \textit{softmax} operation over each row of matrix $\mA$, $\mQ \in \real^{n_q\times d}$ is the matrix containing the $n_q$ query vectors, and $\mK, \mV \in \real^{n \times d}$ are the matrices containing the $n$ key and value vectors respectively, with $d$ being the number of vector dimensions; $\mW^Q$, $\mW^K$, $\mW^V \in \real^{d\times d}$ are the associated weights to perform linear transformations. 

To encode a source sequence, the encoder applies \emph{self-attention}, where $\mQ$, $\mK$ and $\mV$ contain the same vectors coming from the output of the previous layer.\footnote{Initially, $\mQ$, $\mK$, and $\mV$ contain the token embeddings.} 
{
In the decoder, each layer first applies \emph{masked self-attention} over previous-layer states. The resulting vectors are then used as queries to compute \emph{cross-attentions} over the encoder states. For cross-attention, $\mQ$ comprises the decoder self-attention states while $\mK$ and $\mV$ contain the encoder states. The attention mechanism adopted in the  Transformer is considered \textbf{global} since the attention context spans the entire sequence.}


\subsection{Windows in Attentions}

In theory, given enough training data, global attention should be able to model dependencies between the query and the key vectors well. However, in practice we have access to only a limited amount of training data. Several recent studies suggest that incorporating more focused attention over important local regions in the input sequence as an explicit inductive bias could be more beneficial.


In particular, Shaw et al. (\citeyear{Shaw18-relative})  show that adding relative positional biases to the attention scores (Eq. \ref{eq:score})  increases BLEU scores in machine translation. 
Specifically, for each query $\vq_i \in \mQ$ at position $i$ and key $\vk_j \in \mK$ at position $j$, a trainable vector $\va_{i,j}=\vw_{max(-\tau, min(j-i,\tau))}$ is added to the key vector before the query-key dot product is performed. The window size $\tau$ is chosen via tuning. Sperber  et  al.  (\citeyear{Sperber})  also consider local information by restricting self-attention to neighboring representations to improve long-sequence acoustic modeling. Although shown to be effective, their methods only apply to self-attention and not to {cross-attention} where the query vectors come from a different sequence.



That said, Luong et al. (\citeyear{luong2015effective}) are the first to propose a Gaussian-based local attention for \emph{cross-attention}. At each decoding step $t$, their model approximates the source-side pivot position $p_t$ as a function of the decoding state and the source sequence length. Then, local attention is achieved by multiplying the attention score with a \textit{confidence} term derived from a $\mathcal{N}(p_t,\, \sigma^{2})$ distribution. The aligned pivot $p_t$ and the variance $\sigma^{2}$ (a hyper-parameter) respectively represent the center and the size of the local window. 

{Meanwhile, Yang et al. (\citeyear{yang-etal-2018-modeling}) improve the method of Luong et al. (\citeyear{luong2015effective}) by assigning a soft window weight (a Gaussian bias) to obtain a flexible window span.} Despite effective, the  aligned pivot position in the source is determined only by the decoder state, while the encoder states are disregarded - these should arguably give more relevant information regarding the attention spans over the source sequence. Besides, the confidence for local attention span may not strictly follow a normal distribution, but rather vary dynamically depending on the relationship between the query and the key. Furthermore, the approach of Luong et al. (\citeyear{luong2015effective}) is only applicable to {cross-attention} while the one of Yang et al.  (\citeyear{yang-etal-2018-modeling}) works better only for  \emph{encoder} self-attention as shown in their experiments.







Our proposed \emph{differentiable window} approach to local attention addresses the above limitations of previous methods. Specifically, our methods are dynamic and applicable to encoder and decoder self-attentions as well as cross-attention, without any functional constraints. They incorporate encoder states into the local window derivation. They are also invariant to sequence length, which removes the dependence on \textit{global} features from the local context extraction process.

\section{Dynamic Differentiable Window} \label{sec:model}

Our proposed attention method works in two steps: \Ni derive the attention span  for each query vector to attend over, and \Nii compute the respective  attention vector using the  span. In this section, we present our approaches to step \Ni by proposing \emph{trainable soft masking} and \emph{segment-based soft masking}. In the next section, we present our methods to compute the attention vectors. To give the necessary background to understand what can be expected from our method, we first present the \emph{discrete masking} case.

\subsection{Discrete Window Masking}\label{subsec:disc_win_mask}

In this context, we seek to dynamically derive a \emph{boolean mask} vector for each query that will indicate the  window in the key-sequence over which the query should attend. In other words, attentions are only activated on the consecutive positions where the mask vector element is $1$, and the positions with $0$ are canceled out. Let the query vector and the key-sequence be $\vq \in \real^{d}$ and $\mK=(\vk_1, \vk_2, \ldots, \vk_{n})$, respectively. Formally, we  define the local attention mask vector $\vm_q \in \{0,1\}^n$ for the query $\vq$  as follows.

\begin{equation}
    \vm_q^i=\begin{cases}
    1, & \text{if } l_q \le i \le r_q \\
    0, & \text{otherwise}
    \end{cases}
\end{equation}

\noindent where $l_q$ and $r_q$ denote the left and right positional indices that form a \emph{discrete} window $[l_q, r_q]$ over which the query attends. As such, in the standard \emph{global} attention, $l_q=1$ and $r_q=n$ for all the query vectors, and in decoder self-attention, $l_q=1$ and $r_q=t$ for the query vector at decoding step $t$. To facilitate the construction of $\vm_q$, we first define vectors $\phi_k$, $\vf_k$, $\vg_k$ and matrix $\mL_n$ with entries as:

\begin{equation*}
\hspace{-0.6em}\phi^i_k= \begin{cases}
    1, & \text{if } i = k \\
    0, & \text{otherwise}
    \end{cases};
\hspace{0.5em} \vf^i_k= \begin{cases}
    1, & \text{if } i \ge k \\
    0, & \text{otherwise}
    \end{cases}
\hspace{-0.3em}
\end{equation*}
\begin{equation}
\hspace{-0.6em}\vg^i_k= \begin{cases}
    1, & \text{if } i \leq k \\
    0, & \text{otherwise}
    \end{cases};
\hspace{0.5em}\mL_n^{i,j}= \begin{cases}
    1, & \text{if } i \leq j \\
    0, & \text{otherwise}
    \end{cases}
\hspace{-0.8em}
\end{equation}

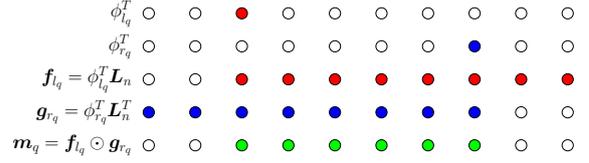
\begin{figure}[t!]
\centering
\scalebox{0.35}{
\begin{tikzpicture}[line cap=round,line join=round, x=1cm,y=1cm]
\draw (-18.5,0) node[anchor=east] {\huge $\phi_{l_q}^T$};
\draw (-18.5,-1.25) node[anchor=east] {\huge $\phi_{r_q}^T$};
\draw (-18.5,-2.5) node[anchor=east] {\huge $\vf_{l_q} = \phi_{l_q}^T \mL_n$};
\draw (-18.5,-3.75) node[anchor=east] {\huge $\vg_{r_q} = \phi_{r_q}^T \mL_n^T$};
\draw (-18.5,-5) node[anchor=east] {\huge $\vm_q = \vf_{l_q} \odot \vg_{r_q}$};
\begin{scriptsize}

\draw [fill=white] (-18,0) circle (6pt);
\draw [fill=white] (-16.25,0) circle (6pt);
\draw [fill=red] (-14.5,0) circle (6pt);
\draw [fill=white] (-12.75,0) circle (6pt);
\draw [fill=white] (-11,0) circle (6pt);
\draw [fill=white] (-9.25,0) circle (6pt);
\draw [fill=white] (-7.5,0) circle (6pt);
\draw [fill=white] (-5.75,0) circle (6pt);
\draw [fill=white] (-4,0) circle (6pt);
\draw [fill=white] (-2.25,0) circle (6pt);

\draw [fill=white] (-18,-1.25) circle (6pt);
\draw [fill=white] (-16.25,-1.25) circle (6pt);
\draw [fill=white] (-14.5,-1.25) circle (6pt);
\draw [fill=white] (-12.75,-1.25) circle (6pt);
\draw [fill=white] (-11,-1.25) circle (6pt);
\draw [fill=white] (-9.25,-1.25) circle (6pt);
\draw [fill=white] (-7.5,-1.25) circle (6pt);
\draw [fill=blue] (-5.75,-1.25) circle (6pt);
\draw [fill=white] (-4,-1.25) circle (6pt);
\draw [fill=white] (-2.25,-1.25) circle (6pt);
\vspace{-0.5em}
\draw [fill=white] (-18,-2.5) circle (6pt);
\draw [fill=white] (-16.25,-2.5) circle (6pt);
\draw [fill=red] (-14.5,-2.5) circle (6pt);
\draw [fill=red] (-12.75,-2.5) circle (6pt);
\draw [fill=red] (-11,-2.5) circle (6pt);
\draw [fill=red] (-9.25,-2.5) circle (6pt);
\draw [fill=red] (-7.5,-2.5) circle (6pt);
\draw [fill=red] (-5.75,-2.5) circle (6pt);
\draw [fill=red] (-4,-2.5) circle (6pt);
\draw [fill=red] (-2.25,-2.5) circle (6pt);
\vspace{-0.5em}
\draw [fill=blue] (-18,-3.75) circle (6pt);
\draw [fill=blue] (-16.25,-3.75) circle (6pt);
\draw [fill=blue] (-14.5,-3.75) circle (6pt);
\draw [fill=blue] (-12.75,-3.75) circle (6pt);
\draw [fill=blue] (-11,-3.75) circle (6pt);
\draw [fill=blue] (-9.25,-3.75) circle (6pt);
\draw [fill=blue] (-7.5,-3.75) circle (6pt);
\draw [fill=blue] (-5.75,-3.75) circle (6pt);
\draw [fill=white] (-4,-3.75) circle (6pt);
\draw [fill=white] (-2.25,-3.75) circle (6pt);
\vspace{-0.5em}
\draw [fill=white] (-18,-5) circle (6pt);
\draw [fill=white] (-16.25,-5) circle (6pt);
\draw [fill=green] (-14.5,-5) circle (6pt);
\draw [fill=green] (-12.75,-5) circle (6pt);
\draw [fill=green] (-11,-5) circle (6pt);
\draw [fill=green] (-9.25,-5) circle (6pt);
\draw [fill=green] (-7.5,-5) circle (6pt);
\draw [fill=green] (-5.75,-5) circle (6pt);
\draw [fill=white] (-4,-5) circle (6pt);
\draw [fill=white] (-2.25,-5) circle (6pt);
\end{scriptsize}
\end{tikzpicture}

}
\caption{Example of $\phi$, $\vf$, and $\vg$ vectors and how the mask vector $\vm_q$ can be derived for $l_q=3$ and $r_q=8$.}
\label{fig:masking_vectors}
\end{figure}

\noindent where $\phi_k \in \{0, 1\}^{n}$ denotes the one-hot representation for a boundary position $k$ (from the left or right of a sequence), and $\vf_k, \vg_k \in \{0, 1\}^n$ are the `rightward' mask vector and `leftward' mask vector, respectively; $\mL_n \in \{0, 1\}^{n \times n}$ denotes a unit-value (1) upper-triangular matrix with $i$ and $j$ being the row and column indices respectively. Figure \ref{fig:masking_vectors} visualizes how these entities appear. Specifically, $\vf_k$ has entry values of $1$'s for position $k$ and its right positions, while $\vg_k$ has entry values of $1$'s for position $k$ and its left positions. As such, $\vf_k$ and $\vg_k$ can be derived from $\phi_k$ and $\mL_n$ as follows.

\begin{eqnarray}
\vf_k = \phi_k^T \mL_n; \hspace{1em} \vg_k = \phi_k^T \mL_n^T 
\end{eqnarray}

\noindent Note that $\vf_k$ can be interpreted as the \emph{cumulative sum} across $\phi_k$, while $\vg_k$ as the \textit{inverse cumulative sum} across $\phi_k$.

Given the above definitions, the mask vector $\vm_q$ for a query $\vq$ to attend over the window $[l_q, r_q]$ in the key sequence such that $1\leq l_q \leq r_q \leq n$ can be achieved by:

\begin{equation}
\vm_q = \vf_{l_q} \odot \vg_{r_q} = (\phi_{l_q}^T \mL_n) \odot (\phi_{r_q}^T \mL_n^T) \label{eqn:masking_vector}
\end{equation}

\noindent where $\odot$ denotes element-wise multiplication. As shown in Figure \ref{fig:masking_vectors}, $\vm_q$ represents the intersection between $\vf_{l_q}$ and $\vg_{r_q}$, and forms a masking span for the attention.


\subsection{Trainable Soft Masking}\label{subsec:trainable_soft_masking}

The above masking method is non-differentiable as $\phi$ is discrete, which makes it unsuitable in an end-to-end neural architecture. In our trainable soft masking method, we approximate the discrete one-hot vector $\phi$ with a pointing mechanism \cite{pointer-net}.\footnote{{However, unlike the standard pointer network, in our case there is no direct supervision for learning the pointing function. Our network instead learns it from the end prediction task.}} Specifically, given the query $\vq$ and the key-sequence $\mK$ as before, we define confidence vectors $\hat{\phi}_{l_q}, \hat{\phi}_{r_q} \in \real^{n}$ as follows.

\begin{eqnarray}
\hat{\phi}_{l_q} &=& \softmax(\frac{\vq^T \mW_L^Q(\mK \mW_L^K)^T}{\sqrt{d}})\label{eqn:soft_phi_lq}\\
\hat{\phi}_{r_q} &=& \softmax(\frac{\vq^T \mW_R^Q(\mK \mW_R^K)^T}{\sqrt{d}})\label{eqn:soft_phi_rq}
\end{eqnarray}

\noindent where $\softmax$ is the \emph{softmax} function as defined before, and $\mW_L^Q, \mW_L^K, \mW_R^Q, \mW_R^K \in \real^{d \times d}$ are trainable parameters. Eq. \ref{eqn:soft_phi_lq}-\ref{eqn:soft_phi_rq} approximate the left and right boundary positions of the mask vector for the query $\vq$. However, contrary to the discrete case, they do not enforce \emph{absolute} cancellation or activation of attention weights on any position in the key-sequence. Instead, they assign a confidence score to each position. This allows the model to gradually correct itself from invalid assignments. Moreover, the \textit{softmax} operations enable differentiability while maintaining the gradient flow in an end-to-end neural architecture.

Note however that the left and right boundary concepts have now become ambiguous since the positions $l_q = \argmax(\hat{\phi}_{l_q})$ and $r_q = \argmax(\hat{\phi}_{r_q})$ are not guaranteed to conform to the constraint $l_q \leq r_q$. {To understand its implication, lets first consider the discrete case in Eq. \ref{eqn:masking_vector}; the element-wise multiplication between $\vf_{l_q}$ and $\vg_{r_q}$ results in a \textit{zero} vector for $\vm_q$ if $l_q > r_q$, canceling out the attention scores entirely.} Although not absolute \textit{zeros}, in the continuous case, $\vm_q$ would potentially contain significantly small values, which renders the attention implausible. To address this, we compute the \emph{soft} mask vector $\hat{\vm}_q$ as follows.

\begin{equation}
\hat{\vm}_q = (\hat{\phi}_{l_q}^T \mL_n) \odot (\hat{\phi}_{r_q}^T \mL_n^T) + (\hat{\phi}_{r_q}^T \mL_n) \odot (\hat{\phi}_{l_q}^T \mL_n^T)\label{eqn:m_q_soft}
\end{equation}

\noindent This formulation has two additive terms; the former constructs the mask vector when $l_q \leq r_q$, whereas the latter is activated when $l_q > r_q$. This ensures a non-zero result regardless of $l_q$ and $r_q$ values. {It can be shown that the values in $\hat{\vm}_q$ represent the expected value of the discrete flags in ${\vm}_q$, i.e., $\hat{\vm}_q$ = $\mathbb{E}({\vm}_q)$; see Appendix for a proof.}

We concatenate the mask vectors horizontally for all the query vectors in $\mQ \in \real^{m \times d}$ to get the mask matrix $\mM \in \real^{m\times n}$. Since the pointing mechanism is invariant to sequence length, the computation of the mask vectors enjoys the same advantages, enabling our models to efficiently perform attentions on any arbitrarily long sequences. In addition, the method is applicable to all attention scenarios -- from decoder to encoder cross-attention, encoder self-attention, and decoder self-attention.

\subsection{Segment-Based Soft Masking}\label{subsec:segment_based_masking}

The soft masking introduced above modulates the attention weight on each token separately which may result in unsmooth attention weights on neighbouring tokens. However, words in a sentence are related and they often appear in chunks or phrases, contributing to a shared meaning. Thus, it may be beneficial to assign identical mask values to the tokens within a segment so that they are {equally treated in the window selection method.} In this section, we propose a novel extension to our soft masking method that {enables the mask vector to share the same masking values for the tokens within a segment in a key-sequence.}

{The main idea is to divide the key-sequence $\mK=(\vk_1, \vk_2, \ldots, \vk_n)$ into $\lceil n / b \rceil$ consecutive segments and to assign the same masking value to the tokens in a segment. The segment size $b$ is considered a hyper-parameter.} We compute the segment-based mask vector $\vm_q'$ similarly as in Eq. \ref{eqn:m_q_soft}, but with $\mL_n$ replaced by $\mJ_n \in \real^{n\times n}$ defined as follows.

\begin{align}
\mJ_n^{i,j} = \begin{cases}
    1, & \text{if } i \leq b\lceil \frac{j}{b} \rceil \\
    0, & \text{otherwise}
    \end{cases}
    \label{eqn:segmentj}
\end{align}

\begin{equation}
\vm_q' = (\hat{\phi}_{l_q}^T \mJ_n) \odot (\hat{\phi}_{r_q}^T \mJ_n^T) + (\hat{\phi}_{r_q}^T \mJ_n) \odot (\hat{\phi}_{l_q}^T \mJ_n^T) \label{eqn:m_q_segment}
\end{equation}

\begin{figure}[t!]
\begin{center}
\includegraphics[width=0.45\textwidth]{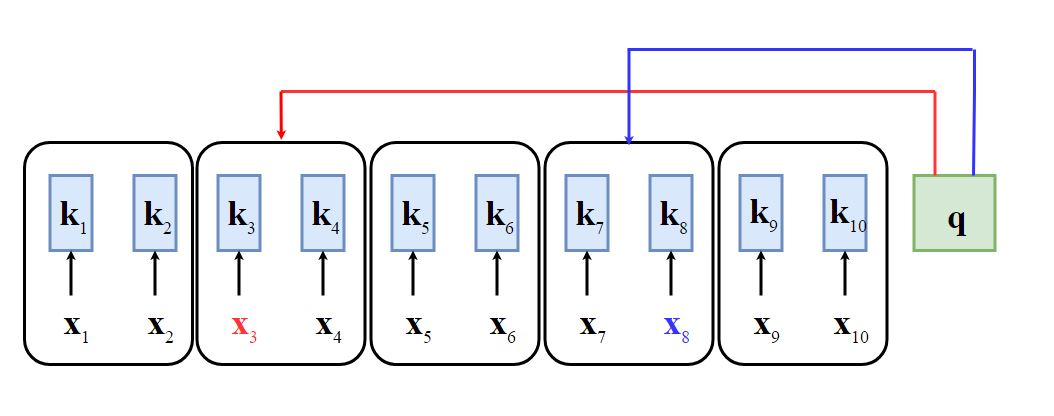}
\end{center}
\vspace{-0.5em}
\caption{Segment-based masking for segment size $=2$. Instead of pointing to the left and right indices of the \emph{tokens}, the soft segment-based method {(approximately)} points to the left and right boundaries of the \emph{segments}, respectively.} 
\label{fig:estimate_left_right_block}
\end{figure}

Eq. \ref{eqn:segmentj} - \ref{eqn:m_q_segment} ensure that all the items in a segment share the same masking value, which is the cumulative sum of the confidence scores in $\hat{\phi}_{l_q}$ and $\hat{\phi}_{r_q}$. For instance, suppose $\hat{\phi}_{l_q}=(a_1,a_2,a_3,\ldots,a_n)$ and segment size $b=2$, then the term $\hat{\phi}_{l_q}^T \mJ_n$ evaluates to
$(\sum_{i=1}^2 a_i, \sum_{i=1}^2 a_i,\sum_{i=1}^4 a_i,\ldots )$,
and $\hat{\phi}_{l_q}^T \mJ_n^T$ evaluates to $(\sum_{i=1}^n a_i, \sum_{i=1}^n a_i,\sum_{i=3}^n a_i,\ldots)$. {Similarly, $\hat{\phi}_{r_q}^T \mJ_n^T$ and $\hat{\phi}_{r_q}^T \mJ_n$ will have segment-level  effects on the cumulative sums. Figure \ref{fig:estimate_left_right_block} visualizes the method with an example for $b=2$.}

One advantage of this approach is that it allows us to control the masking behavior (by varying $b$) without increasing the number of parameters compared to the token-based masking. We also show its effectiveness in our experiments.

\section{Dynamic Window Attention Methods} \label{sec:attn}

Having presented our method to compute the mask vector that defines the attention spans, we now present our methods to incorporate the mask vectors into the attention layers. 

\subsection{Multiplicative Window Attention}

In this approach, the attention weights (Eq. \ref{eq:att}) are (element-wise) multiplied by the mask matrix $\mM$ to confine their attention scope defined by the mask. Formally, the attention scores and outputs are defined as follows.

\begin{eqnarray}
\hspace{-1em} \text{score} \hspace{-0.7em} &=& \hspace{-0.7em} \frac{(\mQ\mW^Q)(\mK\mW^K)^T}{\sqrt{d}}\label{eqn:soft-local:score}\\
\hspace{-2em} \text{att}_{\text{\sc{MW}}} \hspace{-0.7em} &=& \hspace{-0.7em} (\softmax(\text{score}) \odot \mM)(\mV \mW^V) \label{eqn:soft-local:att}
\end{eqnarray}
     
\noindent In this approach, the standard global attention weights are suppressed and partially overshadowed by the attention window imposed by $\mM$. Thus, it can be interpreted as a \textit{local attention} method similar to Luong et al. (\citeyear{luong2015effective}). However, instead of using a static Gaussian bias, we use a dynamic mask to modulate the attention weights. 

\subsection{Additive Window Attention}

Having a \textit{local} attention window could be beneficial, but it does not rule out the necessity of global attention, which has been shown effective in many applications \cite{vaswani2017attention,devlin2018bert}. Thus, we also propose an \emph{additive window attention}, which implements a combination of global attention and local attention. The attention output in this method is formally defined as

\begin{eqnarray}
\hspace{-1em}s_{\text{{glb}}} &=& (\mQ \mW^Q_{\text{glb}})(\mQ\mW^K_{\text{glb}})^T\\
\hspace{-1em}s_{\text{{loc}}} &=& (\mQ \mW^Q_{\text{loc}})(\mQ\mW^K_{\text{loc}})^T \odot \mM \label{eqn:loc-sc}\\
\hspace{-1em}\text{score}_{\text{{AW}}} &=& \frac{s_{glb} + s_{loc}}{\sqrt{d}}\\
\hspace{-1em}\text{att}_{\text{{AW}}} &=& \softmax(\text{score}_{\text{{AW}}})(\mV \mW^V)\label{eqn:global-local:att}
\end{eqnarray}

\noindent where $\mW^Q_{\text{glb}}, \mW^K_{\text{glb}}, \mW^Q_{\text{loc}}$, and $\mW^K_{\text{loc}} \in \real^{d \times d}$ are the weight matrices for global and local attentions.

Compared to the multiplicative window attention where the mask  re-evaluates the global attention weights, additive window attention applies the mask vector to the \textit{local attention scores} ($s_{\text{{loc}}}$), which is then added to the \textit{global attention scores} ($s_{\text{glb}}$) before passing it through the {softmax} function. In this way, the mask-defined local window does not suppress the global context but rather complements it with a local context. Moreover, the resulting attention weights add up to \textit{one}, which avoids attention weights diminishment that could occur in the multiplicative window attention. Additive merger of global and local window components may also facilitate more stable gradient flows.

\subsection{Implementation in Transformer}\label{subsec:impl_transformer}

We now describe how the proposed {dynamic window  attention} methods can be integrated into the Transformer.

\paragraph{Encoder, Decoder and Cross Attentions.} Our proposed methods can be readily applied to the any of the attention layers in the Transformer framework. We could also  selectively apply our methods to different layers in the encoder and decoder. In our initial experiments {on WMT'14 English-German development set,} we observed that the following settings provide more promising performance gains.\\
First, encoder self-attention layers benefit most from \emph{additive window attention}, while {decoder self-attention} layers prefer \emph{multiplicative attention}. This shows that the \textit{global} attention component is more useful when the key sequence is provided entirely in the encoder, while less useful when only the fragmented key sequence (past keys) is visible in the decoder. Second, the above argument is further reinforced as we found that {cross-attention} layers also prefer \emph{additive window attention}, where the entire source sequence is available. Third, cross-attention works better with \emph{segment-based masking}, which  provides smoothness and facilitates phrase (n-gram) based translations.

\paragraph{Lower-layer Local Attentions.} {It has been shown that deep neural models learn simple word features and local syntax in the lower layers, while higher layers learn more complex context-dependent aspects of word semantics. Belinkov et al. (\citeyear{belinkov-etal-2017-neural}) show this on NMT models, while Peters et al. (\citeyear{peters-etal-2018-dissecting}) and Jawahar et al. (\citeyear{jawahar-etal-2019-bert}) show this on representation learning with ELMo and BERT respectively. In other words, local contextual information can still be derived in higher layers with the standard global attention. As such, we propose to apply our dynamic window attention methods only to the first 3 layers of the Transformer network, leaving the top 3 layers intact. Our diverse experiments in the following section support this setup as it offers substantial improvements, whereas using local attention in higher layers does not show gains, but rather increases model parameters.}

\section{Experiment} \label{sec:experiments}
In this section, we present the training settings, experimental results and analysis of our models in comparison with the baselines on machine translation (MT), sentiment analysis, subject verb agreement and language modeling (LM) tasks.

\subsection{Machine Translation}\label{subsec:result:mt}

We trained our models on the standard WMT'16 English-German (En-De) and WMT'14 English-French (En-Fr) datasets containing about 4.5 and 36 million sentence pairs, respectively. For {validation (development)} purposes, we used \emph{newstest2013} for En-De  and a random split from the training set for En-Fr. All translation tasks were evaluated against their respective \emph{newstest2014} test sets, in case-sensitive tokenized BLEU.
We used \emph{byte-pair encoding} \cite{sennrich2015neural} with shared source-target vocabularies of 32,768 and 40,000 sub-words for En-De and En-Fr translation tasks, respectively. We compare our models with three strong baselines: \Ni Transformer Base \cite{vaswani2017attention}, \Nii Transformer Base with Relative Position \cite{Shaw18-relative}, and \Nii Transformer Base with Localness Modeling \cite{yang-etal-2018-modeling}. To ensure a fair comparison, we trained our models and the baselines with the following training setup.

\paragraph{Training Setup.} We followed model specifications in \cite{vaswani2017attention} and optimization settings in \cite{scaling_nmt_ott2018scaling}, with some minor modifications.
Specifically, we used word embeddings of dimension 512, feedforward layers with inner dimension 2048, and multi-headed attentions with 8 heads. 

We trained our models on a single physical GPU but replicated the {8-GPU setup} following the \emph{gradient aggregation} method proposed by Ott et al. (\citeyear{scaling_nmt_ott2018scaling}).

We trained the models for 200,000 \textit{updates} for  En-De and 150,000 \textit{updates} for En-Fr translation tasks. Finally, we averaged the last 5 checkpoints to obtain the final models for evaluation. The segment size $b$ in the segment-based masking method was set to 5.\footnote{We did not tune $b$; tuning $b$ might improve the results further.}

\paragraph{Translation Results.} {We report our translation results in Table \ref{table:avg-bleu-wmt}; \textbf{Enc(AW)} indicates the use of \emph{additive window} (AW) attention in the encoder, \textbf{Dec(MW)} indicates the use of \emph{multiplicative window} (MW) attention in the decoder, and \textbf{Cr(AW,Seg)} indicates the use of additive window attention with \emph{segment-based masking} for cross-attention. The attention module that is not specified in our naming convention uses the default token-based global attention in the Transformer. For example, \textbf{Enc(AW)-Dec(MW)} refers to the model that uses AW attention in the encoder, MW attention in the decoder and the default global attention for cross attention.} 

{We notice that despite a minor increase in the number of parameters, applying our attentions in the encoder and decoder offers about 0.7 and 1.0 BLEU improvements in En-De and En-Fr translation tasks respectively, compared to the Transformer base \cite{vaswani2017attention}. Our model with the segment-based additive method for \textit{cross} attention achieves a similar performance. We observe further improvements as we apply our attentions in all the attention modules of the Transformer. Specifically, our model {Enc(AW)-Cr(AW,Seg)-Dec(MW)} achieves 28.25 and 40.32 BLEU in En-De and En-Fr translation tasks, outperforming Transformer base with localness \cite{yang-etal-2018-modeling} by 0.63 and 0.85 BLEU, respectively.}

\begin{table}[t]
\begin{center}
\resizebox{0.98\columnwidth}{!}{%
\begin{tabular}{lccc} 
\toprule
{\bf Model}     & {\bf \#-params}    & {\bf En-De} & {\bf En-Fr} \\
\midrule
\citet{vaswani2017attention}     &63M    &27.46    & 39.21   \\
\citet{Shaw18-relative}          &63M    &27.56    & 39.37 \\
\citet{yang-etal-2018-modeling}  &63M    &27.62    & 39.47  \\
\midrule
\textbf{Our Models} \\
Enc(AW)-Dec(MW)            &68M     &28.11 & 40.24\\
Cr(AW, Seg)       &65M     &28.13 & 40.06\\
Enc(AW)-Cr(AW,Seg)-Dec(MW)        &73M     &\textbf{28.25} & \textbf{40.32}\\
\bottomrule
\end{tabular}
}
\caption{BLEU scores for different models in WMT'14 English-German  and English-French translation tasks.}
\label{table:avg-bleu-wmt}
\end{center}
\end{table}

\subsection{Ablation Study} {To verify our modeling decisions, we performed an ablation study in the WMT'14 En-De translation task. In particular, we evaluated \Ni the impact of applying our differentiable window attentions in all layers vs. only in certain lower layers of the Transformer network, \Nii which window attention methods (additive or multiplicative) are suitable particularly for the encoder/decoder self-attention and cross-attention, and \Niii the impact of segment-based masking in different attention modules. \Niv training efficiency and performance of our best model with the similar models.} Plus, to further interpret our window-based attention, we also provide the local window visualization.

\begin{table}[t!]
\begin{center}
\resizebox{0.9\columnwidth}{!}{%
\begin{tabular}{lccc} 
\toprule
{\bf Method}    & {\bf Module}    & {\bf Full (6 layers)} & {\bf Partial (3 layers)} \\    
\midrule
Transformer        & -               & 27.46   & -\\
\midrule
AW              & Encoder         & 27.77   & 27.90 \\
MW              & Encoder         & 27.25   & 27.40 \\
\midrule
AW              & Decoder         & 27.73   & 27.85 \\
MW              & Decoder         & 27.88   & 28.04 \\
\midrule
AW              & Cross           & 27.78   & 27.97 \\
MW              & Cross           & 27.58   & 27.79\\
\bottomrule
\end{tabular}
}
\caption{Evaluation of {Additive Window (AW)} and {Multiplicative Window (MW)} attentions in encoder/decoder self attention and cross attention for full vs. partial settings.}
\label{table:wmt_ablation_full_vs_partial}
\end{center}
\end{table}

\paragraph{Full vs. Partial.} {Table \ref{table:wmt_ablation_full_vs_partial} shows BLEU scores for the Transformer models that employ our window-based attentions in all 6 layers (\textbf{Full}) vs. only in the first 3 layers (\textbf{Partial}), as well as  the methods used in different attention modules (encoder/decoder self-attention, cross-attention). We can see that almost all the models with window-based methods in the first 3 layers outperform those that use them in all 6 layers.} This gives the setup significant advantages as it performs not only better in BLEU but also requires less parameters. 

The results also show that multiplicative window ({MW}) attention is preferred in decoder self-attention, while additive window ({AW}) is more suitable for encoder self-attention and for cross-attention. This suggests that the global context, which is maintained in AW, is more useful when it is entirely available like in encoder self-attention and cross attention. In contrast,  incomplete and partially-generated context in decoder self-attention may induce more noise than information, where MW attention renders better performance than AW.

\begin{table}[t!]
\begin{center}
\resizebox{0.92\columnwidth}{!}{%
\begin{tabular}{lcc} 
\toprule
{\bf Model}     & {\bf Token-based} & {\bf Segment-based} \\

\midrule
Cr(AW) & 27.97& 28.13  \\
Enc(AW)-Dec(MW) & 28.11 & 27.91\\
\bottomrule
\end{tabular}
}
\caption{BLEU scores for token- and segment-based masking in cross attention and encoder self-attention. The decoder self-attention always uses token-based masking.}
\label{table:wmt_ablation_segment}
\end{center}
\end{table}

\begin{figure*}[t!]
\begin{subfigure}{.35\linewidth}
  \centering
  \includegraphics[width=0.90\textwidth]{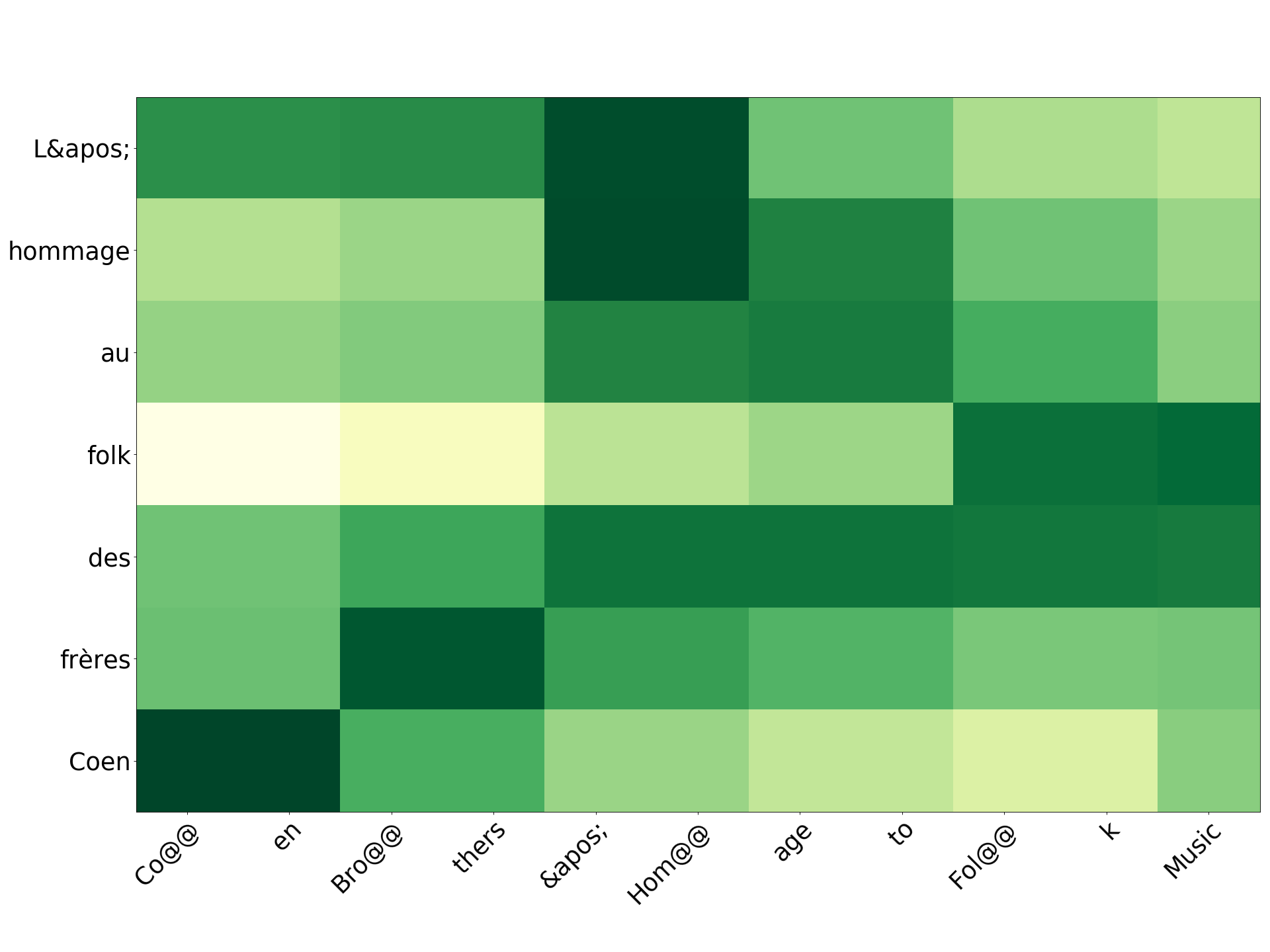}
    \caption{Local masking scores ($\mM$).}
  \label{fig:local_window_scores}
\end{subfigure}\hspace*{-1em}
\begin{subfigure}{.35\linewidth}
  \centering
    \includegraphics[width=0.90\textwidth]{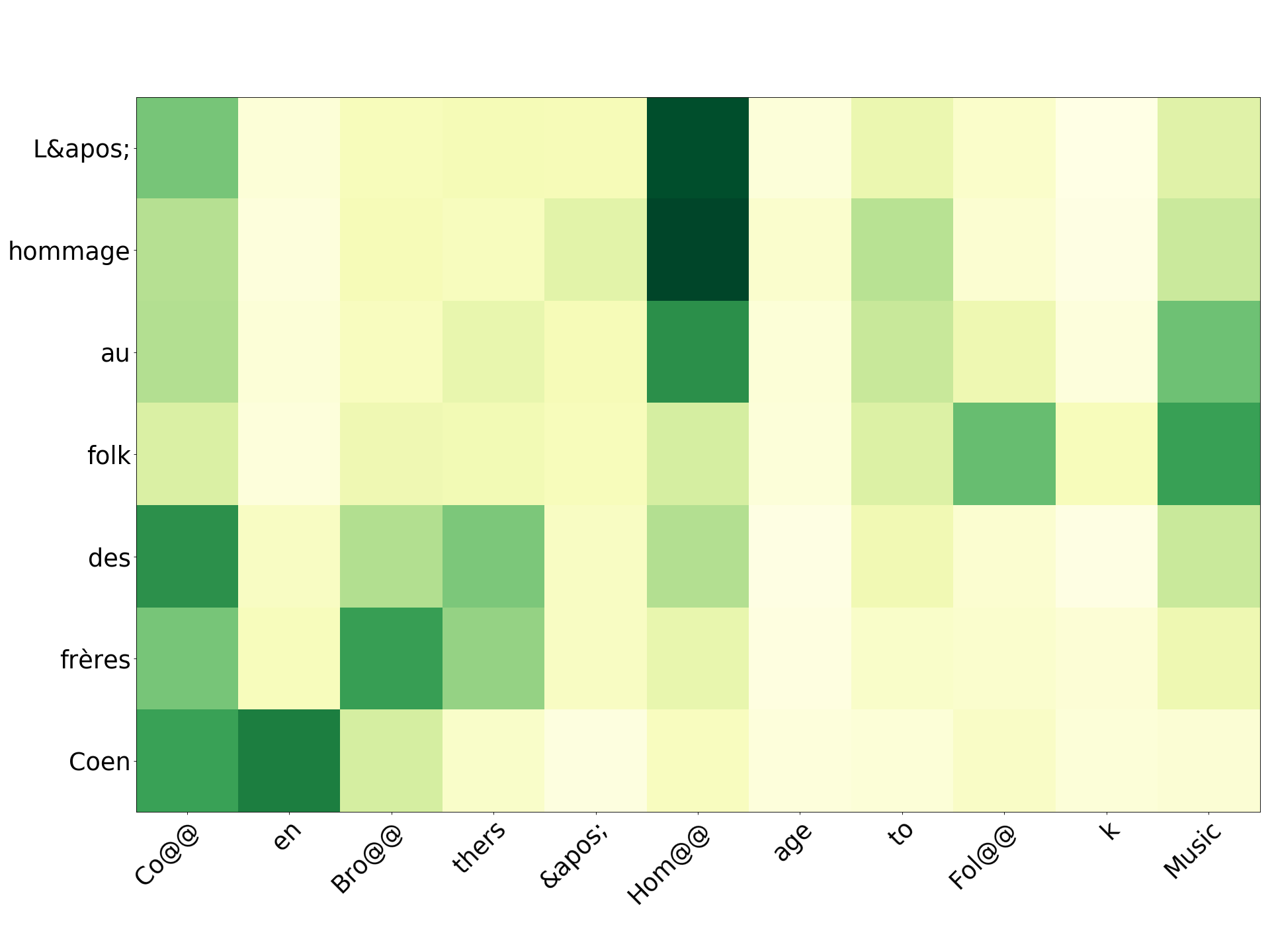}
\caption{Our attention scores.}
  \label{fig:model_attention_scores}
\end{subfigure}\hspace*{-1em}
\begin{subfigure}{.35\linewidth}
  \centering
    \includegraphics[width=0.90\textwidth]{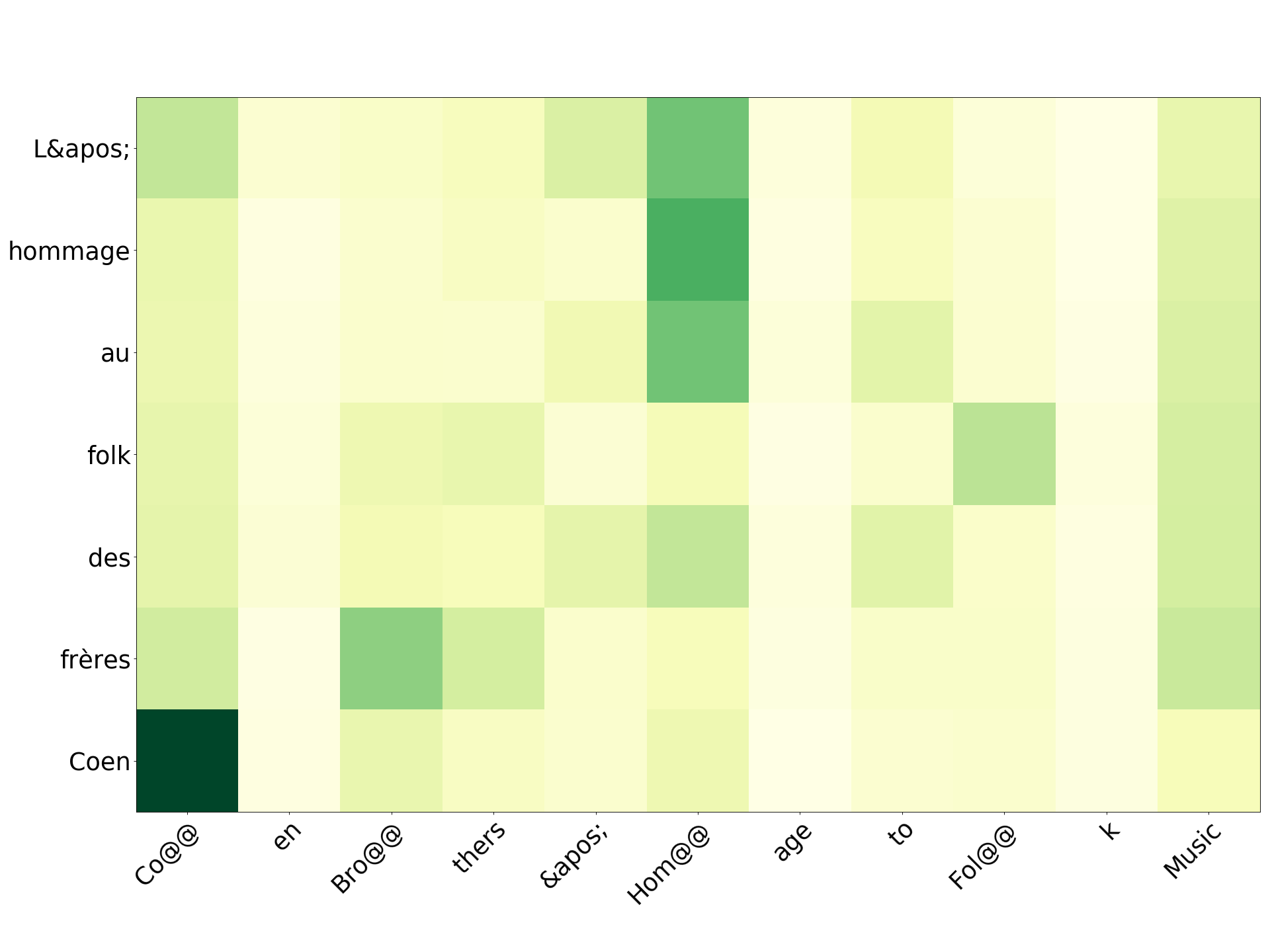}
  \caption{Transformer attention scores.}
  \label{fig:normal_attention_scores}
\end{subfigure}
\caption{Visualization of masking scores, and attention scores for our and the original Transformer models.}
\label{fig:global_local_scores}

\end{figure*}

\paragraph{Token- vs. Segment-based.} Table \ref{table:wmt_ablation_segment} compares the results for using {token-based} vs. {segment-based} masking methods in different attention modules of the network. Note that it is preferred for {decoder self-attention} to adopt token-based masking since the decoder cannot point to \textit{unfinished} segments in autoregressive generation, {if it had used segment-based masking.} We see  that segment-based additive window masking   outdoes its token-based counterpart (28.13 vs. 27.97 BLEU) for {cross-attention}. Meanwhile, for {encoder self-attention}, token-based masking performs better than segment-based masking by 0.2 BLEU. This suggests that segments (or phrases) represent better \textit{translation} units than tokens, justifying its performance superiority in cross-lingual attention but not in mono-lingual (self-attention) encoding. 

\begin{table}[t]
\centering
\resizebox{0.98\columnwidth}{!}{%
\begin{tabular}{lccc}
    \bf Model & \bf \#-params & \bf \# steps/sec &  \bf  BLEU \\ 
    \midrule
    \citet{vaswani2017attention}   &63M &   1.20 & 27.46  \\ 
    \citet{yang-etal-2018-modeling} &63M  &1.07  &27.62      \\
    \citet{vaswani2017attention} 7 layers &69M  &   1.05 & 27.74  \\ 
    \citet{vaswani2017attention} 8 layers &75M  &   0.99 & 27.89  \\ 
    \midrule
    Enc(AW)-Cr(AW,Seg)-Dec(MW)        &73M &1.04    &\textbf{28.25}\\
    \bottomrule
  \end{tabular}
  }
  \caption{Training efficiency and size of similar models}
  \label{table:speed_model_size_compare}
\end{table}

\paragraph{Speed and Parameters.} As shown in table \ref{table:speed_model_size_compare}, our training efficiency is competitive to the baselines. That is, the training speed for our model is $1.04$ steps/sec which is similar to \citet{yang-etal-2018-modeling}. Besides, our model outperforms the Transformer with 8 layers, which has more parameters. This suggests that our performance gain may not come from additional parameters, but rather from a better inductive bias through the dynamic window attention.

\paragraph{Local Window Visualization.} 

To further interpret our window-based attentions, Figure \ref{fig:local_window_scores} shows the cross-attention soft masking values ($\hat{\vm}_q$) on the source tokens for each target token in an En-Fr test sample assigned by our Enc(AW)-Cr(AW,Seg)-Dec(MW) model. The darker the score, the higher the attention is from a target token to a source token. We can see the relevant subwords are captured by the attentions quite well, which promotes ngram-level alignments. For instance, the mask ($\hat{\vm}_q$) guides the model to evenly distribute attention scores on sub-words ``Co@@'' and  ``en'' (Fig. \ref{fig:model_attention_scores}), while standard attention is biased towards ``Co@@'' (Fig. \ref{fig:normal_attention_scores}). Similar phenomenon can be seen for ``Bro@@'' and ``thers'' (towards ``fr\`{e}res'').

\subsection{Text Classification}\label{subsec:result:sentiment_analysis}

We evaluate our models on the Stanford Sentiment Treebank (SST) \cite{D13-1170},  IMDB sentiment analysis \cite{maas-EtAl:2011:ACL-HLT2011} and Subject-Verb Aggreement (SVA) \cite{TACL972} tasks. We compare our attention methods (incorporated into the Transformer encoder) with the encoders of Vaswani et al. (  \citeyear{vaswani2017attention}), Shaw et al.  (\citeyear{Shaw18-relative}) and Yang et al. (\citeyear{yang-etal-2018-modeling}).

\paragraph{Training Setup.} As the datasets are quite small compared to the MT datasets, we used \textit{tiny versions} of our models as well as the baselines.\footnote{As specified in https://github.com/tensorflow/tensor2tensor.} Specifically, the models consist of a 2-layer Transformer {encoder} with 4 attention heads, 128 hidden dimensions and 512 feedforward inner dimensions. In these experiments, our attention methods are applied only to the first layer of the network. We trained for 3,000, 10,000 and 10,000 updates for SST, IMDB and SVA tasks, respectively on a single GPU machine.


 
 \begin{table}[t]
\begin{center}
\resizebox{0.95\columnwidth}{!}{%
\begin{tabular}{lccc} 
\toprule
{\bf Model}                  & {\bf STT} & {\bf IMDB} &{\bf SVA}\\
\midrule
\citet{vaswani2017attention}           & 79.36 & 83.65 & 94.48     \\
\citet{Shaw18-relative} & 79.73 & 84.61  & 95.27\\
\citet{yang-etal-2018-modeling} & 79.24 & 84.13 & 95.00\\
\midrule
Enc (MW)            & 79.70 & 85.09  &95.95\\
Enc (AW)        & \textbf{82.13}    & \textbf{87.98}  &\textbf{96.19} \\

\bottomrule
\end{tabular}
}
\caption{Classification accuracy on Stanford Sentiment Treebank (SST) and IMDB sentiment analysis and Subject-Verb Agreement(SVA) tasks.}
\label{table:sentiment_sva}
\end{center}
\end{table}
 
\paragraph{Results.} 
Table \ref{table:sentiment_sva} shows the results. Our multiplicative window approach ({Enc (MW)}) achieves up to 79.7\%, 85.1\% and 95.95\% accuracy in SST, IMDB and SVA, exceeding Transformer   \cite{vaswani2017attention} by 0.4\%, 1.35\% and 1.47\%, respectively. Our additive window attention ({Enc (AW)}) renders even more improvements. Specifically, it outperforms Transformer with relative position (Shaw et al. \citeyear{Shaw18-relative}) by 2.4\% and 3.37\%, 0.92\% reaching 82.13\%, 87.98\% and 96.19\% accuracy in SST, IMDB and SVA, respectively. In fact, the results demonstrate consistent trends with our earlier MT experiments: additive window attention outdoes its multiplicative counterpart in the encoder, where the entire key sequence is available. 


\subsection{Language Modeling}\label{subsec:result:lm}

Finally, to demonstrate our proposed methods as effective general purpose NLP components, we evaluate them on the One Billion Word LM Benchmark dataset \cite{Chelba-LM}.
The dataset contains 768 million words of data compiled from WMT 2011 News Crawl data, with a vocabulary of 32,000 words. We used its held-out data as the test set.

\paragraph{Training Setup.} As the LM dataset is considerably large, we used the same model settings as adopted in our MT experiments.  For these experiments, we only trained the models on virtually 4 GPUs for 100,000 \textit{updates} using gradient aggregation on a single GPU machine. 
Note that only the self-attention based autoregressive {decoder} of the Transformer framework is used in this task. Therefore, the method of Yang et al. (\citeyear{yang-etal-2018-modeling}) is not applicable to this task.

\begin{table}[t]
\begin{center}
\resizebox{0.65\columnwidth}{!}{%
\begin{tabular}{lc} 
\toprule
{\bf Model}                  & {\bf Perplexity}  \\
\midrule
\citet{vaswani2017attention}  & 46.37       \\
\citet{Shaw18-relative}       & 46.13\\
\midrule
Dec (MW)            & \textbf{44.00}         \\
Dec (AW)            & 44.95         \\

\bottomrule
\end{tabular}
}
\caption{Perplexity scores on 1-billion-word language modeling benchmark (the lower the better).}
\label{table:lm}
\end{center}
\end{table}

\paragraph{Results.} 
Table \ref{table:lm} shows the {perplexity} scores. As can be seen, our multiplicative and additive window attention models both surpass Transformer \cite{vaswani2017attention} by 2.37 and 1.42 points respectively, reaching 44.00 and 44.95 perplexity scores respectively.
In addition, it is noteworthy that similar to MT experiments, multiplicative attention  outperforms the additive one on this task, where the decoder is used. 
This further reinforces the claim that where the global context is not \textit{fully} available like in the decoder, the incomplete global context may induce noises into the model. Thus, it is effective to embrace dynamic local window attention to suppress the global context, for which the  multiplicative window attention is designed.

\section{Conclusion}\label{sec:conclusion}

We have presented a novel Differential Window method for dynamic window selection, and used it to improve the standard attention modules by enabling more focused attentions. Specifically, we proposed Trainable Soft Masking and Segment-based Masking, which can be applied to encoder/decoder self-attentions and cross attention. 

We evaluated our models on four NLP tasks including machine translation, sentiment analysis, subject verb agreement and language modeling. Our experiments show that our proposed methods outperform the baselines significantly across all the tasks. All in all, we demonstrate the benefit of incorporating the differentiable window in the attention. In the future, we would like to extend our work to make a syntactically-aware window that can automatically learn tree (or phrase) structures.

\section*{Acknowledgments}

We would like to express our gratitude to Yi Tay and our anonymous reviewers for their insightful feedback on our paper. Shafiq Joty would like to
thank the funding support from his Start-up Grant
(M4082038.020).

\bibliography{reference}
\bibliographystyle{acl_natbib}
\newpage
\section*{Appendix} \label{sec:appen}
\subsection*{Proof: $\hat{\vm}_q = \mathbb{E}({\vm}_q)$}

The probability of left and right boundary for a query $q$: 

\begin{eqnarray}
\hat{\phi}_{l_q} &=& \softmax(\frac{\vq^T \mW_L^Q(\mK \mW_L^K)^T}{\sqrt{d}})\\
\hat{\phi}_{r_q} &=& \softmax(\frac{\vq^T \mW_R^Q(\mK \mW_R^K)^T}{\sqrt{d}})
\end{eqnarray}

For any $k$,  

\begin{eqnarray}
p(f_k = 1) = p(l_q \le k) = \sum_{\hat{\phi}_{l_q} \le k} \hat{\phi}_{l_q} = (\hat{\phi}_{l_q}^T \mL_n)_k \\
p(g_k = 1) = p(r_q \ge k) = \sum_{\hat{\phi}_{r_q} \ge k} \hat{\phi}_{r_q} = (\hat{\phi}_{r_q}^T \mL_n^T)_k
\end{eqnarray}

Since $f_k$ and $g_k$ are binary values, 

\begin{eqnarray}
\hat{f}_k = p(f_k = 1) = \mathbb{E}(f_k)\\
\hat{g}_k = p(g_k = 1) = \mathbb{E}(g_k)
\end{eqnarray}

Hence, 

\begin{equation}
\hat{\vm}_q = \hat{\vf}_{l_q} \odot \hat{\vg}_{r_q} + \hat{\vf}_{r_q} \odot \hat{\vg}_{l_q} = \mathbb{E} (\vm_q) 
\end{equation}

\end{document}